\documentclass[11pt,a4paper]{article}
\usepackage{arabtex}
\usepackage{utf8}
\usepackage[hyperref]{eacl2021}

\usepackage{times}
\usepackage{latexsym}
\usepackage{times}
\usepackage{url}
\usepackage{latexsym}
\usepackage{amsmath}
\usepackage{booktabs}
\usepackage{graphicx}
\graphicspath{ {./images/} }
\usepackage{url}
\usepackage{adjustbox,multirow}
\usepackage{tabularx}
\usepackage{amsmath}
\usepackage{pgf}
\usepackage{color,soul}
\usepackage{float}
\usepackage[ruled,vlined, linesnumbered]{algorithm2e}
\usepackage{textcomp}
\usepackage{fixltx2e}

\usepackage[toc,page]{appendix}
\usepackage{subfig}
\usepackage{float}
\usepackage{xcolor}
\usepackage{comment}
\usepackage{textcomp}
\usepackage{microtype}

\usepackage{tablefootnote}
\usepackage[hide]{boxnotes}
\newcommand{\eat}[1]{} 
\newcommand{\mam}[1]{\textcolor{purple}{#1}}

\iftrue 
  \renewcommand{\mam}[1]{#1}
\fi

\aclfinalcopy 

\newcommand*{\affaddr}[1]{#1} 
\newcommand*{\affmark}[1][*]{\textsuperscript{#1}}
\newcommand*{\email}[1]{\texttt{#1}}

\title{Self-Training Pre-Trained Language Models for Zero- and Few-Shot Multi-Dialectal Arabic Sequence Labeling}

 \author{%
Muhammad Khalifa \affmark[1], Muhammad Abdul-Mageed\affmark[2], 
Khaled Shaalan\affmark[3]\\
\affaddr{\affmark[1]Department of Computer Science, Cairo University}\\
\affaddr{\affmark[2]Natural Language Processing Lab, University of British Columbia}\\
\affaddr{\affmark[3]The British University in Dubai}\\
\email{m.khalifa@grad.fci-cu.edu.eg} \\ \email{muhammad.mageed@ubc.ca,}
\email{khaled.shaalan@buid.ac.ae}
}

\begin{document}
\maketitle
\setcode{utf8}

\begin{abstract}
A sufficient amount of annotated data is usually required to fine-tune pre-trained language models for downstream tasks. Unfortunately, attaining labeled data can be costly, especially for multiple language varieties and dialects. 
We propose to self-train pre-trained language models in zero- and few-shot scenarios to improve performance on data-scarce varieties using only resources from data-rich ones. We demonstrate the utility of our approach in the context of Arabic sequence labeling by using a language model fine-tuned on Modern Standard Arabic (MSA) only to predict named entities (NE) and part-of-speech (POS) tags on several dialectal Arabic (DA) varieties. We show that self-training is indeed powerful, improving zero-shot MSA-to-DA transfer by as large as \texttildelow 10\% F$_1$ (NER) and 2\% accuracy (POS tagging). We acquire even better performance in few-shot scenarios with limited amounts of labeled data. We conduct an ablation study and show that the performance boost observed directly results from the unlabeled DA examples used for self-training. Our work opens up opportunities for developing DA models exploiting only MSA resources and it can be extended to other languages and tasks.\footnote{Our code and fine-tuned models can be accessed at: \url{https://github.com/mohammadKhalifa/zero-shot-arabic-dialects}}
\end{abstract}



\section{Introduction}
\label{intro}

%
%
    

Neural language models~\cite{xu2000can,bengio2003neural} with vectorized word representations~\cite{mikolov2013efficient} are currently core to a very wide variety of NLP tasks. In specific, using representations from transformer-based~\cite{vaswani2017attention} language models~\cite{devlin2018bert,liu2019roberta}, pre-trained on large amounts of unlabeled data and then fine-tuned on labeled task-specific data, has become a popular approach for improving downstream task performance. This pre-training then fine-tuning scheme has been successfully applied to several tasks, including question answering \cite{yang2019enhancing}, social meaning detection~\cite{mageedL2020:aranet}, text classification~\cite{liu2019roberta}, named entity recognition (NER), and part-of-speech (POS) tagging~\cite{tsai2019small,conneau2019unsupervised}. The same setup also works well for cross-lingual learning~\cite{lample2019cross,conneau2019unsupervised}.


Given that it is very expensive to glean labeled resources for all language varieties and dialects, a question arises: ``How can we leverage resource-rich dialects to develop models nuanced to downstream tasks for resource-scarce ones?". In this work, we aim to answer this particular question by
applying self-training to unlabeled target dialect data. 
  We empirically show that self-training is indeed an effective strategy in \textit{zero-shot} (where no gold dialectal data are included in training set, Section ~\ref{sec:msa-da-zero}) and \textit{few-shot} (where a given number of gold dialectal data points is included in training split, Section~\ref{sec:few-shot}). 

Our few-shot experiments reveal that self-training is always a useful strategy that \textit{consistently} improves over mere fine-tuning, even when \textit{all} dialect-specific gold data are used for fine-tuning. In order to understand why this is the case (i.e., why combining self-training with fine-tuning yields better results than mere fine-tuning), we perform an extensive error analysis based on our NER data. We discover that self-training helps the model most (\% = 59.7) with improving false positives. This includes DA tokens whose MSA orthographic counterparts \cite{shaalan2014survey}  are either named entities or trigger words that frequently co-occur with named entities in MSA. Interestingly, such out-of-MSA tokens occur in highly dialectal contexts (e.g., interjections and idiomatic expressions employed in interpersonal social media communication) or ones where the social media context in which the language (DA) is employed affords more freedom of speech~\cite{alshehri2020osact4} and a platform for political satire. We present our error analysis in Section~\ref{sec:error}.

\textbf{Context:} Language use in social media tends to diverge from `standard', offline norms~\cite{danet2007multilingual,herring2015computer}. For example, users employ slang, emojis, abbreviations, letter repetitions, and other types of playful practices. This poses a challenge for processing social media data in general. However, there are other challenges specific to Arabic that motivate our work. More specifically, we choose Arabic to apply our approach since it affords a rich context of linguistic variation: In addition to the standard variety, MSA, Arabic also has several spoken dialects~\cite{abdul2018you,bouamor2019madar,mageedetal2020nadi,mageed2020micro}, which differ significantly from the written MSA \cite{zaidan2014arabic} thus offering an excellent context for studying our problem. Arabic dialects differ among themselves and from MSA at various linguistic levels: lexical, phonological, morphological, and syntactic. This makes our case much more challenging than that of standard vs. social media English, for example. For a good zero-shot performance in our case, a model is required to accommodate not only lexical distance between MSA and DA, but also differences in word formation and syntax (related to POS tags, for example) and lexical ambiguity (as the meaning of the same token can vary cross-dialectically). This makes the zero-shot setting even harder, where the performance drops ~20\% F1 points (See section ~\ref{sec:msa-da-zero}).

From a geopolitical perspective, Arabic also has a strategic significance. This is a function of Arabic being the native tongue of $~$ 400 million speakers in 22 countries, spanning across two continents (Africa and Asia)\footnote{https://www.internetworldstats.com/stats19.htm}. In addition, the three dialects of our choice, namely Egyptian (EGY), Gulf (GLF), and Levantine (LEV), are popular dialects that are widely used online. This makes our resulting models highly useful in practical situations at scale. Pragmatically, ability to develop NLP systems on dialectal tasks with no-to-small labeled dialect data immediately eases a serious bottleneck. Arabic dialects differ among themselves and from MSA at all linguistic levels, posing challenges to traditional NLP approaches. 
We also note that our method is language-independent, and we hypothesize it can be directly applied to other varieties of Arabic or in other linguistic contexts for other languages and varieties.

\textbf{Tasks:} We apply our methods on two sequence labeling tasks, where we have access to both MSA and DA gold data. In particular, as mentioned above, we perform experiments on POS tagging and NER. Each of these tasks has become an integral part of various other NLP applications, including question answering, aspect-based sentiment analysis, machine translation, and summarization, and hence our developed models should have wide practical use. Again, we note that our approach itself is task-independent. The same approach can thus be applied to other tasks involving DA. We leave testing our approach on other languages, varieties, and tasks for future research.  

\newblock

\textbf{Contributions:} Our work offers the following contributions:
\begin{enumerate}
    \item We study the problem of MSA-to-DA transfer in the context of sequence labeling and show that when training on MSA data only, a wide performance gap exists between testing on MSA and DA. That is, models fine-tuned on MSA generalize poorly to DA in zero-shot settings.
    
    \item We propose self-training to improve zero- and few-shot MSA-to-DA transfer. Our approach requires little-to-no labeled DA data. We evaluate extensively on 3 different dialects, and show that our method indeed narrows the performance gap between MSA and DA by a margin as wide as \texttildelow 10\% F$_1$ points.
    
    \item We develop state-of-the-art models for the two sequence labeling tasks (NER and POS).
\end{enumerate}

We now introduce our method.

\section{Method}~\label{sec:method}
\begin{figure}[t!] 
\hspace{0.6cm}
\includegraphics[width=6.5cm, height=5.5cm]{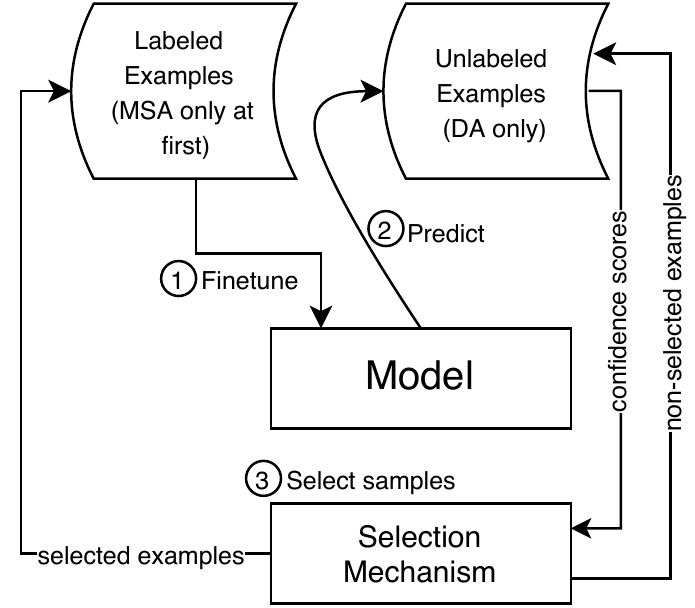}
\caption{MSA-to-DA Self-training transfer.}
\label{Fig:Model}
\end{figure}

While the majority of labeled Arabic datasets are in MSA, most daily communication in the Arab world is carried out in DA. In this work, we show that models trained on MSA for NER and POS tagging generalize poorly to dialect inputs when used in zero-shot-settings (i.e., no dialect data used during training). Across the two tasks, we test how self-training would fare as an approach to leverage unlabeled DA data to improve performance on DA. As for self-training, it involves training a model using its own predictions on a set of unlabeled data identical from its original training split. Our proposed self-training procedure is given two sets of examples: a labeled set $L$ and an unlabeled set $U$. To perform zero-shot MSA-to-DA transfer, MSA examples are used as the labeled set, while unlabeled DA examples are the unlabeled set. As shown in Figure ~\ref{Fig:Model}, each iteration of the self-training algorithm consists mainly of three steps. First, a pre-trained language model is fine-tuned on the labeled MSA examples $L$. Second, for every unlabeled DA example $u_i$, we use the model to tag each of its tokens to obtain a set of predictions and confidence scores for each token $p_{u_i} = (l^{(i)}_{1}, c^{(i)}_{1}), (l^{(i)}_{2}, c^{(i)}_{2}), ... (l^{(i)}_{|u_i|}, c^{(i)}_{|u_i|})$, where $(l^{(i)}_{j}, c^{(i)}_{j})$ are the label and confidence score (softmax probability) for the $j$-th token in $u_i$. Third, we employ a selection mechanism to identify examples from $U$ that are going to be added to $L$ for the next iteration.

For a selection mechanism, we experiment with both a thresholding approach and a fixed-size~\cite{dong2019robust} approach. In the thresholding method, a threshold $\tau$ is applied on the minimum confidence per example. That is, we only add an example $u_i$ to $L$ if $\min \limits_{(l^{(i)}_{j}, c^{(i)}_{j}) \in p_{u_i}}{c^{(i)}_j} \geq \tau$. \mam{See Algorithm~\ref{alg}}. The fixed-size approach involves, at each iteration, the selection of the top $S$ examples with respect to the minimum confidence score $\min \limits_{(l^{(i)}_{j}, c^{(i)}_{j}) \in p_{u_i}}{c^{(i)}_j}$ , where $S$ is a hyper-parameter. We experiment with both approaches and report results in Section~\ref{sec:res}. 

\begin{algorithm}[ht]
\footnotesize
\SetAlgoLined
\textbf{Given} set $L$ of labeled MSA examples, set $U$ of unlabeled DA examples, $\tau$ parameter for probability threshold selection.

\textbf{repeat}

  \Indp Fine-tune model $M$ for $K$ epochs on labeled MSA examples $L$;\\
  
  \For{$u_i \in U$}{
  Obtain prediction $p_{u_i}$ on unlabeled DA example $u_i$ using model $M$;\\
  \uIf{$\min \limits_{(l^{(i)}_{j}, c^{(i)}_{j}) \in p_{u_i}}{c^{(i)}_j} \geq \tau$}{  remove $u_i$ from $U$ and add it to $L$;\\
  }
  }
  \Indm
  \textbf{until} stopping criterion satisfied
 \caption{\label{alg}MSA-to-DA Self-Training}
\end{algorithm}

For our language model, we use XLM-RoBERTa~\cite{conneau2019unsupervised}, XML-R for short. XLM-R is a cross-lingual model, and we choose it since it is reported to perform better than the multilingual mBERT ~\cite{devlin2018bert}.
XLM-R also uses Common Crawl for training, which is more likely to have dialectal data than the Arabic Wikipedia (used in mBERT), making it more suited to our work. We now introduce our experiments.

\section{Experiments}\label{sec:exps}

We begin our experiments with evaluating the standard fine-tuning performance of XLM-R models on both NER and POS tagging against strong baselines. We then use our best models from this first round to investigate the MSA-to-DA zero-shot transfer, showing a significant performance drop even when using pre-trained XLM-R. Consequently, we employ self-training for both NER and POS tagging in zero- and few-shot settings, showing substantial performance improvements in both cases. We now introduce our datasets.


\subsection{Datasets}~\label{subsec:data}
\noindent 
\textbf{NER:} For our work on NER, we use 4 datasets: ANERCorp~\cite{Benajiba2007}, \texttildelow 150K tokens; ACE 2003 ~\cite{mitchell2003tides} BNews (BN-2003), \texttildelow 15K tokens; ACE 2003 Newswire (NW-2003), \texttildelow 27K tokens; and Twitter~\cite{darwish2013named}, \texttildelow 81K tokens. Named entity types in all datasets are \textit{location (LOC)}, \textit{organization (ORG)}, and \textit{person (PER)}.

\textbf{POS Tagging:} There are a number of Arabic POS tagging datasets, mostly on MSA~\cite{maamouri2004penn} but also on dialects such as EGY~\cite{maamouri2014developing}. To show that the proposed approach is able to work across multiple dialects, we ideally needed data from more than one dialect. Hence, we use the multi-dialectal dataset from~\cite{darwish2018multi}, comprising 350 tweets from each of the 4 varieties MSA, EGY, GLF and LEV. This dataset has 21 POS tags, some of which are suited to social media (since it is derived from Twitter). We show the POS tag set from ~\cite{darwish2018multi} in Table~\ref{Tab:pos-tags} in Appendix ~\ref{sec:pos-tagset}. We now introduce our baselines.


\subsection{Baselines}
For the \textbf{NER task}, we use the following baselines:





\begin{itemize}
    \item \textbf{NERA}\textbf{~\cite{abdallah2012integrating}}: A hybrid system of rule-based features and a decision tree classifier.
    \item \textbf{WC-BiLSTM}\textbf{~\cite{gridach2016character}}: A character- and a word-level Bi-LSTM with a conditional random fields (CRF) layer. 
    \item \textbf{WC-CNN}\textbf{~\cite{khalifa2019character}}: A character- and a word-level CNN with  a CRF layer.
    \item \textbf{mBERT}\textbf{~\cite{devlin2018bert}}: A fine-tuned multilingual BERT-Base-Cased (110M parameters), pre-trained with a masked language modeling objective on the Wikipedia corpus of 104 languages (including Arabic). For fine-tuning, we find that (based on experiments on our development set) a learning rate of $6 \times 10^{-5}$ works best with a dropout of 0.1. 
\end{itemize}
In addition, we compare to the published results in~\cite{shaalan2014hybrid}, AraBERT~\cite{antoun2020arabert}, and CAMel~\cite{obeid2020camel} for the ANERCorp dataset. We also compare to the published results in~\cite{khalifa2019character} for the 4 datasets.

For the \textbf{POS tagging task}, we compare to our own implementation of WC-BiLSTM (since there is no published research that uses this method on the task, as far as we know) and run mBERT on our data. We also compare to the CRF results published by~\cite{darwish2018multi}. In addition, for the Gulf dialect, we compare to the BiLSTM with compositional character representation and word representations (CC2W+W) published results in~\cite{alharbi2018part}.

\subsection{Experimental Setup}

Our main models are XLM-RoBERTa base architecture XLM-R\textsubscript{B} $(L=12, H=768, A=12, \textnormal{ 270M params})$ and  XLM-RoBERTa large architecture XLM-R\textsubscript{L} $(L=24, H=1024, A=16,\textnormal{ 550M params})$, where $L$ is number of layers, $H$ is the hidden size, $A$ is the number of self-attention heads. For XLM-R experiments, we use Adam optimizer with $1e^{-5}$ learning rate, batch size of 16. We typically fine-tune for 20 epochs, keeping the best model on the development set for testing. We report results on the test split for each dataset, across the two tasks. For all BiLSTM experiments, we use the same hyper-parameters as~\cite{khalifa2019character}.

For the standard fine-tuning experiments, we use the same train/development/test split as in \cite{khalifa2019character} for NER, and the same split provided by \cite{darwish2018multi} for POS tagging. 
For all the self-training experiments, we use the dialect subset of the Arabic online news commentary (AOC) dataset~\cite{zaidan2011arabic}, comprising the EGY, GLF, and LEV varieties limiting to equal sizes of 9K examples per dialect (total =27K)~\footnote{We note that our approach could be scaled with an even bigger unlabeled dataset, given the performance gains we report with self-training in this work.}. We use the split from~
\cite{elaraby2018deep} of AOC, removing the dialect labels and just using the comments themselves for our self-training. Each iteration involved fine-tuning the model for $K=5$ epochs. As a stopping criterion, we use early stopping with patience of 10 epochs. Other hyper-parameters are set as listed before. For selecting confident samples,  we experiment with a fixed number of top samples $S=[50, 100, 200]$ and selection based on a probability threshold $\tau=[0.80, 0.90, 0.95]$ (softmax values)~\footnote{It is worth noting that our $S$ values are similar to those used in~\cite{dong2019robust}. We also experimented with other values for $\tau$ and $S$, but found them sub-optimal and hence we report performance only for the listed values of these two hyper-parameters here.}. For all evaluations, we use the \textit{seqeval} toolkit.\footnote{\url{https://github.com/chakki-works/seqeval}.} 


\section{Results}\label{sec:res}
\subsection{Fine-tuning XLM-R}\label{sec:finetune}

Here, We show the resuts of standard fine-tuning of XLM-R for the two tasks in question. We start by showing the result of fine-tuning XLM-R on the \textbf{named entity task}, on each of the 4 Arabic NER (ANER) datasets listed in Section~\ref{subsec:data}. Table~\ref{Tab:ner} shows the test set macro F$_1$ score on each of the 4 ANER datasets.  Clearly, the fine-tuned XLM-R models outperform other baselines on all datasets, except on the NW-2003 where WC-CNN~\cite{khalifa2019character} performs slightly better than XLM-R\textsubscript{L}.

For \textbf{POS Tagging}, Table ~\ref{Tab:pos} shows test set word accuracy  of the XLM-R models compared to baselines. Again, XLM-R models (both base and large) outperform all other models. A question arises why XLM-R models outperform both mBERT and AraBERT. As noted before, for XLM-R vs. mBERT, XLM-R was trained on much larger data: CommonCrawl for XLM-R vs. Wikipedia for mBERT. Hence, the \textit{larger dataset} of XLM-R is giving it an advantage over mBERT. For comparison with AraBERT, although the pre-training data for XLM-R and AraBERT may be comparable, even the smaller XLM-R model (XLM-R\textsubscript{B}) has more than twice the number of parameters of the BERT\textsubscript{BASE} architecture on which AraBERT and mBERT are built (270M v. 110M). Hence, XLM-R model \textit{capacity} gives it another advantage. We now report our experiments with zero-shot transfer from MSA to DA.

\begin{table*}[h]
\begin{center}
\footnotesize
\begin{tabular}{lllll}
\toprule
\bf Model &  \bf ANERCorp &  \bf BN-2003 & \bf NW-2003 &  \bf Twitter \\ \hline
NERA \cite{abdallah2012integrating} & 88.77 & ~ ~ --  & ~ ~ --  &  ~ ~ --    \\
CAMeL \cite{obeid2020camel} & 85.00 & ~ ~ --  & ~ ~ --  &  ~ ~ --      \\
Hybrid \cite{shaalan2014hybrid} & 90.66 & ~ ~ --  & ~ ~ --  &  ~ ~ --      \\
WC-BiLSTM \cite{gridach2016character}  & 88.56  & 94.92 & 90.32 & 64.93   \\
WC-CNN \cite{khalifa2019character} & 88.77 & 94.12 & \bf 91.20 & 65.34   \\
mBERT (ours) & 85.86 & 89.52  &  87.19  & 58.92 \\
AraBERT \cite{antoun2020arabert} & 84.2 & ~ ~ --  & ~ ~ --  &  ~ ~ --     \\

\hline
XLM-R\textsubscript{B} (ours) & 87.75 & 95.35 & 85.25 & 60.39   \\
XLM-R\textsubscript{L} (ours) &  \textbf{91.43} & \textbf{97.33} & 91.10 &  \textbf{68.91}   \\

\bottomrule
\end{tabular}
\end{center}
\caption{\label{Tab:ner} Test set macro F$_1$ scores for NER.}
\end{table*}

\begin{table*}[h]
\begin{center}
\footnotesize
\begin{tabular}{lllll}
\toprule
\bf Model &  \bf   \bf MSA & \bf EGY & \bf GLF & \bf LEV  \\ \hline
BiLSTM (CC2W + W) \cite{alharbi2018part}  & ~~--  &  ~~--  & 89.7 & ~~--  \\
CRF \cite{darwish2018multi}    &  93.6  & 92.9 & 87.8 & 87.9 \\
WC-BiLSTM  (ours)   & 94.63  & 93.41 & 88.79 & 86.13\\
mBERT (ours)  & 90.57  &  92.88  & 87.85 & 72.30  \\

XLM-R\textsubscript{B} (ours)  & 96.30 & 94.70 & 92.18 & 89.98\\
XLM-R\textsubscript{L} (ours)  & \bf 98.21 & \bf 97.00 & \bf 94.41 & \bf 93.19\\
\bottomrule
\end{tabular}
\end{center}
\caption{\label{Tab:pos} Test set accuracy for POS Tagging.}
\end{table*}

\subsection{MSA-DA Zero-Shot Transfer}\label{sec:msa-da-zero}

We start by the discussion of \textbf{NER experiments}. Since there is no publicly available purely dialectal NER dataset on which we can study MSA-to-DA transfer, we needed to find DA data to evaluate on. We observed that the dataset from ~\cite{darwish2013named} contains both MSA and DA examples (tweets). Hence, we train a binary classifier\footnote{The model we use is XLM-R\textsubscript{B} fine-tuned on the AOC using~\cite{elaraby2018deep} split. We achieve development and test accuracies of 90.3\% and 89.4 \%, respectively, outperforming the best results in \cite{elaraby2018deep}.}
to distinguish DA data from MSA. We then extract examples that are labeled with probability $p > 0.90$ as either DA or MSA. We obtain 2,027 MSA examples (\textit{henceforth}, \texttt{Twitter-MSA}) and 1,695 DA examples (\textit{henceforth}, \texttt{Twitter-DA}), respectively. We split these into development and test sets with 30\% and 70\% ratios. As \textbf{for POS Tagging}, we already have the three previously used DA datasets, namely EGY, GLF and LEV. We use those for the zero-shot setting by omitting their training sets and using only the development and test sets.


We first study how well models trained for NER and POS tagging on MSA data only will generalize to DA inputs during test time. We evaluate this zero-shot performance on both the XLM-R\textsubscript{B} and XLM-R\textsubscript{L} models. \textbf{For NER}, we train on ANERCorp (which is pure MSA) and evaluate on both Twitter-MSA and Twitter-DA. While for POS tagging, we train on the MSA subset~\cite{darwish2018multi} and evaluate on the corresponding test set for each dialect. As shown in Table ~\ref{Tab:zero-shot-all}, for NER, a significant generalization gap of around 20 \% F$_1$ points exists between evaluation on MSA and DA using both models. While for \textbf{POS tagging}, the gap is as large as 18.13 \% accuracy for the LEV dialect with XLM-R\textsubscript{B}. The smallest generalization gap is on the GLF variety, which is perhaps due to the high overlap between GLF and MSA~\cite{alharbi2018part}. In the next section, we evaluate the ability of self-training to close this MSA-DA performance gap.

\begin{table*}[h]
\begin{center}
\footnotesize
\begin{tabular}{c|c|c|c|c|c|c}
\toprule

\textbf{Model}
    & \multicolumn{2}{c|}{\textbf{NER}}
        &  \multicolumn{4}{c}{\textbf{POS}} \\

\hline
 & \textbf{MSA} & \textbf{DA} & \textbf{MSA} &  \textbf{EGY} & \textbf{GLF} & \textbf{LEV} \\
\hline

XLM-R\textsubscript{B} & 60.42 & 40.07 & 96.30 & 78.38 & 83.72 & 78.17 \\
XLM-R\textsubscript{L} & 68.32 & 47.35 & 98.21 & 82.28 & 85.95 & 81.24 \\
\bottomrule
\end{tabular}
\end{center}
\caption{Zero-shot transfer results on DA For NER (macro F$_1$) and POS Tagging (accuracy). Models are trained on MSA only and evaluated on DA. Datasets used are: Twitter-MSA and Twitter-DA \citep{darwish2013named} for NER, and Multi-dialectal \citep{darwish2018multi} for POS tagging.
\label{Tab:zero-shot-all} 
}
\end{table*}

\subsection{Zero-shot Self-Training}\label{sec:msa-da-zero-st}
Here, \textbf{for NER}, similar to Section~\ref{sec:msa-da-zero}, we train on  ANERCorp (pure MSA) and evaluate on Twitter-MSA and Twitter-DA. Table ~\ref{Tab:st-ner} shows self-training NER results employing the selection mechanisms listed in Section~\ref{sec:method}, and with different values for $S$ and $\tau$. The best improvement is achieved with the thresholding selection mechanism with a $\tau=0.90$, where we have an F$_1$ gain of 10.03 points. More generally, self-training improves zero-shot performance in all cases albeit with different F$_1$ gains. 
It is noteworthy, however, that the much higher-capacity large model deteriorates on MSA if self-trained (dropping from 68.32\% to 67.21\%). This shows the ability of the large model to learn representations very specific to DA when self-trained. It is also interesting to see that the best self-trained base model achieved 50.10\% F$_1$, outperforming the large model before the latter is self-trained (47.35\% in the zero-shot setting). As such, we conclude that \textit{\textbf{a base self-trained model, with less computational capacity, can (and in our case does) improve over a large (not-self-trained) model} that needs significant computation}. The fact that, when self-trained, the large model improves 15.35\% points over the base model in the zero-shot setting (55.42 vs. 40.07) is remarkable.


\begin{table}[h]
\begin{center}
\footnotesize
\begin{tabular}{l|l|l}
\toprule
\bf Model & \bf MSA& \bf DA \\ \hline
XLM-R\textsubscript{B} & 61.88 & 40.07 \\ 

XLM-R\textsubscript{B}, ST, S=50 & 60.98  &  43.88 \\
XLM-R\textsubscript{B}, ST, S=100 & 61.13 &  42.01 \\
XLM-R\textsubscript{B}, ST, S=200 & 61.46 &  43.49 \\

XLM-R\textsubscript{B}, ST, $\tau=0.80$ & \textbf{63.36} &  46.97 \\
XLM-R\textsubscript{B}, ST, $\tau=0.90$ & 61.02 & \bf 50.10 \\
XLM-R\textsubscript{B}, ST, $\tau=0.95$ & 62.25 & 47.91 \\
\hline
XLM-R\textsubscript{L} & \textbf{68.32} & 47.35 \\
XLM-R\textsubscript{L} + ST, $\tau=0.90$ & 67.21 & \textbf{55.42} \\

\bottomrule
\end{tabular}
\end{center}
\caption{\label{Tab:st-ner} Zero-short self-training (ST) NER results. Models trained on  ANERCorp (pure MSA) and evaluated on Twitter-MSA and Twitter-DA we extract from~\cite{darwish2018multi}. Self-training boosts the performance on DA data by 10\% macro F1 points with XLM-R\textsubscript{B} and $\tau=0.90$.}
\end{table}

As \textbf{for POS tagging}, we similarly observe consistent improvements in zero-shot transfer with self-training (Table~\ref{Tab:zero}). The best model achieves accuracy gains of 2.41\% (EGY), 1.41\% (GLF), and 1.74\% (LEV). Again, this demonstrates the utility of self-training pre-trained language models on the POS tagging task even in absence of labeled dialectal POS data (zero-shot).

\begin{table}[h]
\begin{center}
\footnotesize
\begin{tabular}{l|l|l|l|l}
\toprule
\bf Model & \bf           MSA & \bf EGY & \bf GLF & \bf LEV \\ \hline
XLM-R\textsubscript{B} & 96.30 & 78.38 & 83.72 & 78.17 \\ 
XLM-R\textsubscript{B}, ST, S=50 & ~~~ -- & \textbf{80.79} & \textbf{85.13} & \textbf{79.91}  \\
XLM-R\textsubscript{B}, ST, S=100 & ~~~ -- & 80.43 & 84.74 & 79.16  \\
XLM-R\textsubscript{B}, ST, S=200 & ~~~ -- & 78.75 & 84.21 & 79.40  \\ 
XLM-R\textsubscript{B}, ST, $\tau$=0.90 & ~~~ -- & 79.52 & 83.97 &  79.21 \\ 
XLM-R\textsubscript{B}, ST, $\tau$=0.85& ~~~ -- & 78.97 & 83.53 &  79.06   \\ 
XLM-R\textsubscript{B}, ST, $\tau$=0.80 & ~~~ -- & 78.88 & 83.72 &  78.50 \\ 
\hline
XLM-R\textsubscript{L} & 98.21 & 82.28 & 85.95 & 81.24 \\
XLM-R\textsubscript{L}, ST,  S=50 &  ~~~ -- & \textbf{82.65} & \textbf{87.76} & \textbf{83.70} \\


\bottomrule
\end{tabular}
\end{center}
\caption{\label{Tab:zero} Zero-shot POS tagging transfer accuracy when training on MSA only. \textbf{ST:} self-training.}
\end{table}

\subsection{Few-Shot Self-Training}\label{sec:few-shot}
\begin{figure*}[t!]
\includegraphics[width=\textwidth, height=3.9cm]{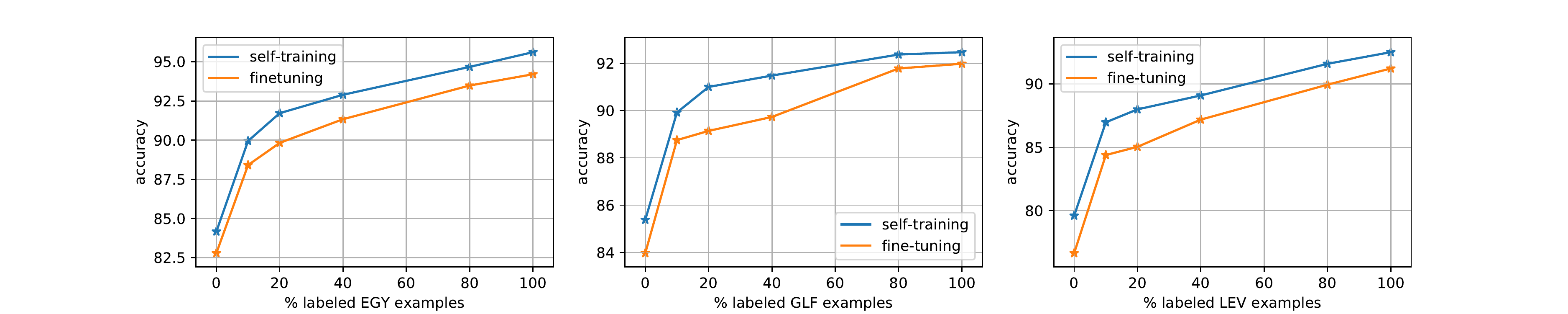}
\caption{Development accuracy as labeled DA data is added to the training MSA data. \textbf{Left:} Results on EGY.  \textbf{Middle: } Results on GLF. \textbf{Right:} Results on LEV. We use fixed-size selection with $S=100$ for self-training models.}
\label{Fig:few-shot}
\end{figure*}

We also investigate whether self-training would be helpful in scenarios where we have access to some gold-labeled DA data (as is the case with POS tagging). Here, we evaluate the few-shot performance of self-training as increasing amounts of predicted DA data are added to the gold training set. This iteration of experiments focuses exclusively on \textbf{POS tagging}, using a fixed-size $S=100$ of predicted cases for self-training and the XLM-R base model. Figure~\ref{Fig:few-shot} shows how POS tagging test accuracy improves as the percentage of gold DA examples added to the MSA training data increases from 0\% to 100\% on the three dialects (EGY, GLF, and LEV). Comparing these results to those acquired via the standard fine-tuning settings without self-training, we find that self-training does \textit{consistently} improve over fine-tuning. This improvement margin is largest with only 20\% of the gold examples.

\subsection{Ablation Study}
Here, we conduct an ablation study with the NER task as our playground in order to verify our hypothesis that the performance boost primarily comes from using unlabeled DA data for self-training. By using a MSA dataset with the same size as our unlabeled DA one\footnote{We use a set of MSA tweets from the AOC dataset mentioned before.}, we can compare the performance of the self-trained model in both settings: MSA and DA unlabeled data. We run 3 different self-training experiments using 3 different values for $\tau$ using each type of unlabeled data. Results are shown in table ~\ref{Tab:st-msa}. While we find slight performance boost due to self-training even with MSA unlabeled data, the average F1 score with unlabeled DA is better by 2.67 points, showing that using unlabeled DA data for self-training has helped the model adapt to DA data during testing.
\begin{table}[h]
\begin{center}
\footnotesize
\begin{tabular}{l|l|l}
\toprule
\bf Setting & \bf Unlabeled MSA& \bf Unlabeled DA \\ \hline
ST, $\tau=0.80$ & 43.88 &  44.46 \\
ST, $\tau=0.90$ & 44.69 & 47.83 \\
ST, $\tau=0.95$ & 43.43 & 46.87 \\
\hline
Avg & 43.67 & \textbf{46.34} \\ 
\bottomrule
\end{tabular}
\end{center}
\caption{\label{Tab:st-msa} Ablation experiment with MSA unlabeled data for zero-shot NER. Development set macro F1 is shown when using both unlabeled MSA and DA data with the same size. Average performance with DA unlabeled data is higher showing the effect of unlabeled DA on model final performance.}
\end{table}

\section{Error analysis}\label{sec:error}



\begin{table}[ht!]
\footnotesize
    \centering

    \begin{tabular}{c|c|c|c}
        \toprule
        \textbf{Measure}  &  \textbf{FT} & \textbf{ST} & \textbf{\% improvement} \\
        \hline
        True Positives & 155 & 165 & 6.5 \% \\
        False Positive & 159 & 64 & \textbf{59.7 \%}\\
        False Negative & 162 & 168 & -3.7 \% \\
        True Negative & 5,940 & 6,035 & 1.5 \% \\
        \bottomrule
    \end{tabular}
    \caption{Comparison of error categories in percentage between the fine-tuned model (FT) and the model combining fine-tuned+self-trained (ST) model, based on the dialectal part of the dev set of the NER task. }
    \label{tab:st-fp}
\end{table}


\begin{table*}[]
    \centering
    \footnotesize 

    \begin{tabular}{l|c|l|c|l|c}
        \toprule
        \textbf{no.} & \textbf{Token} &\textbf{Eng.} &\textbf{MSA} & \textbf{Context/Explanation} & \textbf{FT Pred.} \\
        \hline
        (1) & \<نبي > & we want  & \<نريد>  &  \< نبي نعرف من...>    (\textit{we want to know who}) & PER  \\

       (2) &   \<ماكانوا> & wasn't &  \<لم يكونوا>
          & \<أغلب الي ماكانوا مصدقين>
          (\textit{most of those who wasn't believing}) & LOC  \\

         (3) & \<لوووول > & LOL &  \<ضحك>  & \< لوووول...> (interjection) & PER  \\
        (4) & \< عشان > & for & \<لكي > &\<تبي بطاريات عشان تلعب > (\textit{she wants batteries to play})  & LOC  \\

       (5) &  \<دلوقتي > & now &  \<الآن >  & \<...اقنعوه ينزل دلوقتي>  \textit{(convince him to move now})  &         PER  \\

       (6) &   \<ايش> & what &  \<ماذا>
          & \<ايش رأيك>
          (\textit{what do you think?}) & PER \\

        (7) & \<قادر > & capable &  \<قادر> &\<وبقدرة قادر...>  (\textit{magically}; idiomatic expression)& PER \\



        

          

    
   \bottomrule
    \end{tabular}
    \caption{Sample false positives mitigated by self-training. These were correctly predicted as the unnamed entity ``O" by the self-trained model.}
    \label{tab:fp_analysis}
\end{table*}


To understand why self-training the pre-trained language model, when combined with fine-tuning, improves over mere fine-tuning, we perform an error analysis. For the error analysis, we focus on the NER task where we observe a huge self-training gain. We use the development set of Twitter-DA (See section ~\ref{sec:msa-da-zero-st}) for the error analysis. We compare predictions of the standard fine-tuned XLM-R\textsubscript{B} model (FT) and the best performing self-training ($\tau=0.9$) model (ST) on the data, and provide the confusion matrices of both models with gold labels in Table ~\ref{tab:conf-mat} (in Appendix ~\ref{app-err}). The error analysis leads to an interesting discovery: The greatest benefit from the ST model comes mostly from reducing \textit{false positives} (see Table~\ref{tab:st-fp}). In other words, self-training helps regularize the model predictions such that tokens misclassified by the original FT model as a named entities are now correctly tagged as \textit{unnamed entity} ``O". 

To understand why the ST model improves false positive rate, we manually inspect the cases it correctly identifies that were misclassified by the FT model. We show examples of these cases in Table~\ref{tab:fp_analysis}. As the table shows, the ST model is able to identify dialectal tokens whose equivalent MSA forms can act as trigger words (usually followed by a PER named entity). We refer to this category as \textbf{\textit{false trigger words}}. An example is the word  \<نبي > ``prophet" (row 1 in Table~\ref{tab:fp_analysis}). A similar example that falls within this category is in row (2), where the model is confused by the  token  \<الى> ( ``who" in EGY, but ``to" in MSA and hence the wrong prediction as LOC). A second category of errors is caused by \textbf{\textit{non-standard social media language}}, such as use of letter repetition in interjections (e.g., in row (3) in  Table~\ref{tab:fp_analysis}). In these cases, the FT model also assigns the class PER, but the ST model correctly identifies the tag as ``O". A third class of errors arises as a result of \textit{\textbf{out-of-MSA}} vocabulary. For example, the words in rows (4-6) are all out-of-MSA where the FT model, not knowing these, assigns the most frequent named entity label in train (PER). A fourth category of errors occurs as a result of a token that is usually part of a named entity in MSA, that otherwise functions as part of an \textit{\textbf{idiomatic expression}} in DA. Row (7) in Table~\ref{tab:fp_analysis} illustrates this case. Table ~\ref{tab:fp_analysis_app} in Appendix ~\ref{app-err} provides more examples.


We also investigate errors shared by both the FT and ST models (errors which the ST model also could not fix). Some of these errors result from the fact that often times both MSA and DA use the same word for both person and location names. Row (1) in Table ~\ref{tab:shared-errors} (in Appendix~\ref{append:error}) is an example where the word ``Mubarak", name of the ex-Egypt President, is used as LOC. Other errors include \textit{out-of-MSA} tokens mistaken as named entities. An example is in row (3) in Table ~\ref{tab:shared-errors}, where \<بأمارة> ,(``proof" or  ``basis" in EGY) is confused for \<بإمارة> (``emirate", which is a location). \textit{False trigger words}, mentioned before, also play a role here. An example is in row (7) where \<يابطل> is confused for  PER due to the trigger word \<يا> ``Hey!" that is usually followed by a person name. \textit{\textbf{Spelling mistakes}} cause the third source of errors, as in row (4). We also note that even with self-training, detecting ORG entities is more challenging than PER or LOC. The problem becomes harder when such organizations are not seen in training such as in rows (8) \<الاخوان المسلمين>, (9) \<قناة العربية> and  (10) \<المجلس العسكري>, all of which do not occur in the training set (ANERCorp). 

\textbf{False negatives.} The ``regularizing" effect caused by self-training we discussed thus far can sometimes produce \textit{false negatives} as shown in Table ~\ref{tab:fn-analysis}. We see a number of named entities that were misclassified by the self-trained model as unnamed ones. As an example, we take the last name \<الجنزوري> which was classified both correctly and incorrectly in different contexts by the self-trained model. Context of correct classification is ``\<هاش تاج لكمال الجنزوري> ", while it is ``\<ماسك على الناس كلها سي دي الا الجنزوري  ماسك > \\ \< عليه فلوبي>" for the incorrect classification. First, we note that \<الجنزوري> is not a common name (zero occurrences in the MSA training set). Second, we observe that in the correct case, the word was preceded by the first name \<كمال> which was correctly classified as PER, making it easier for the model to assign PER to the word afterwards as a surname.
\begin{table}[]
    \centering 
     \begin{tabular}{l|c|c|c|c}
        \toprule
        \textbf{no.} & \textbf{Word}  & \textbf{Gold} & \textbf{FT} & \textbf{ST}  \\
        \hline
        (1) & \<الاخوان > &  ORG  &  ORG  & O \\
        (2) &   \<للبرادعي > &   PER &   PER &  O \\
         (3) &  \< مجدي الجلاد > &  PER &   PER  & O \\
        (4) &   \<فان ديزل>  &  PER &   PER  &  O \\
        (5) &  \<الجنزوري> &  PER &   PER  &  O \\
        (6) &  \<زين يسون>   &  PER &   PER  &  O \\
         \bottomrule
    \end{tabular}
    
    \caption{\textbf{NER task.} Sample false negatives produced by self-training.}
    \label{tab:fn-analysis}
\end{table}



\section{Related Work}\label{sec:rel}
\textbf{Sequence Labeling.} Recent work on sequence labeling usually involves using a word- or character-level neural network with a CRF layer ~\cite{lample2016neural,ma2016end}. These architectures have also been applied to Arabic sequence tagging~\cite{gridach2016character,alharbi2018part,khalifa2019character,al2020transfer,el2019arabic}, producing better or comparable results to classical rule-based approaches \cite{shaalan2014hybrid}. We refer the interested reader to \cite{shoufan2015natural} and  \cite{al2018deep} for surveys on Arabic NLP.

\textbf{Pre-trained Language Models.}
Language models, based on Transformers~\cite{vaswani2017attention}, and pre-trained with the masked language modeling (MLM) objective have seen wide use in various NLP tasks. Examples include BERT~\cite{devlin2018bert}, RoBERTa~\cite{liu2019roberta}, MASS~\cite{song2019mass}, and ELECTRA~\cite{clark2020electra}. While they have been applied to several tasks, including text classification, question answering, named entity recognition~\cite{conneau2019unsupervised}, and POS tagging~\cite{tsai2019small}, a sufficiently large amount of labeled data is required for good performance. Concurrent with our work,~\newcite{mageed2020marbert} released MARBERT, a language model trained on a large amount of dialectal Arabic data. However, the extent to which dialect-specific models such as MARBERT can alleviate lack of labeled data remains untested.

\textbf{Cross-lingual Learning.} Cross-lingual learning is of particular importance due to the scarcity of labeled resources in many of the world's languages. The goal is to leverage existing labeled resources in high-resource languages (such as English) to optimize learning for low-resource ones. In our case, we leverage MSA resources for building DA models. With proximity to our work,~\newcite{kim2017cross} trained a POS tagger for different languages using English-resources only using two BiLSTM networks to learn common and language-specific features.  \newcite{xie2018neural} made use of bilingual word embeddings with self-attention to learn cross-lingual NER for low-resource languages.

Multilingual extensions of LMs have emerged through joint pre-training on multiple languages. Examples include mBERT~\cite{devlin2018bert}, XLM~\cite{lample2019cross} and XLM-RoBERTa~\cite{conneau2019unsupervised}. Such multilingual models have become useful for few-shot and zero-shot cross-lingual settings, where there is little or no access to labeled data in the target language. For instance~\newcite{conneau2019unsupervised} evaluated a cross-lingual version of RoBERTa~\cite{liu2019roberta}, namely XLM-R, on cross-lingual learning across different tasks such as question answering, text classification, and named entity recognition. 

\textbf{Self-Training.} Self-Training is a semi-supervised technique to improve learning using unlabeled data. Self-training has been successfully applied to NER~\cite{kozareva2005self}, POS tagging~\cite{wang2007semi}, parsing~\cite{sagae2010self} and text classification~\cite{van2016predicting}. Self-training has also been applied in cross-lingual settings when gold labels are rare in the target language.~\newcite{hajmohammadi2015combination} proposed a combination of active learning and self-training for cross-lingual sentiment classification. ~\newcite{pan2017cross} made use of self-training for named entity tagging and linking across 282 different languages. Lastly,~\newcite{dong2019robust} employed self-training to improve zero-shot cross-lingual classification with mBERT \cite{devlin2018bert}.



\section{Conclusion}\label{sec:conc}
Even though pre-trained language models have improved many NLP tasks, they still need labeled data for fine-tuning. We show how self-training can boost the performance of pre-trained language models in zero- and few-shot settings on various Arabic varieties. We apply our approach to two sequence labeling tasks (NER and POS), establishing new state-of-the-art results on both. Through in-depth error analysis and an ablation study, we uncover why our models work and where they can fail. Our method is \textit{language}- and \textit{task-agnostic}, and we believe it can be applied to other tasks and language settings. We intend to test this claim in future research. Our research also has bearings to ongoing work on language models and self-training, and interactions between these two areas can be the basis of future work.  All our models and code are publicly available.

\section*{Acknowledgements}\label{sec:acknow}
MAM gratefully acknowledges support from the Natural Sciences and Engineering Research Council of Canada, the Social Sciences and Humanities Research Council of Canada, Canadian Foundation for Innovation, Compute Canada (\url{www.computecanada.ca}) and UBC ARC-Sockeye (\url{https://doi.org/10.14288/SOCKEYE}).\\

\bibliography{ref}

\begin{thebibliography}{53}
\expandafter\ifx\csname natexlab\endcsname\relax\def\natexlab#1{#1}\fi

\bibitem[{Abdallah et~al.(2012)Abdallah, Shaalan, and
  Shoaib}]{abdallah2012integrating}
Sherief Abdallah, Khaled Shaalan, and Muhammad Shoaib. 2012.
\newblock Integrating rule-based system with classification for {Arabic} named
  entity recognition.
\newblock In \emph{International Conference on Intelligent Text Processing and
  Computational Linguistics}, pages 311--322. Springer.

\bibitem[{Abdul-Mageed et~al.(2018)Abdul-Mageed, Alhuzali, and
  Elaraby}]{abdul2018you}
Muhammad Abdul-Mageed, Hassan Alhuzali, and Mohamed Elaraby. 2018.
\newblock You tweet what you speak: A city-level dataset of arabic dialects.
\newblock In \emph{Proceedings of the Eleventh International Conference on
  Language Resources and Evaluation (LREC 2018)}.

\bibitem[{Abdul-Mageed et~al.(2020{\natexlab{a}})Abdul-Mageed, Elmadany, and
  Nagoudi}]{mageed2020marbert}
Muhammad Abdul-Mageed, AbdelRahim Elmadany, and El~Moatez~Billah Nagoudi.
  2020{\natexlab{a}}.
\newblock {ARBERT} \& {MARBERT}: Deep bidirectional transformers for {Arabic}.
\newblock \emph{arXiv preprint arXiv:2101.01785}.

\bibitem[{Abdul-Mageed et~al.(2020{\natexlab{b}})Abdul-Mageed, Zhang, Bouamor,
  and Habash}]{mageedetal2020nadi}
Muhammad Abdul-Mageed, Chiyu Zhang, Houda Bouamor, and Nizar Habash.
  2020{\natexlab{b}}.
\newblock Nadi 2020: The first nuanced arabic dialect identification shared
  task.
\newblock In \emph{Proceedings of the Fifth Arabic Natural Language Processing
  Workshop (WANLP 2020)}, pages 97--110, Barcelona, Spain.

\bibitem[{Abdul-Mageed et~al.(2020{\natexlab{c}})Abdul-Mageed, Zhang, Elmadany,
  and Ungar}]{mageed2020micro}
Muhammad Abdul-Mageed, Chiyu Zhang, AbdelRahim Elmadany, and Lyle Ungar.
  2020{\natexlab{c}}.
\newblock Micro-dialect identification in diaglossic and code-switched
  environments.
\newblock In \emph{Proceedings of the 2020 Conference on Empirical Methods in
  Natural Language Processing (EMNLP)}, pages 5855--5876.

\bibitem[{Abdul-Mageed et~al.(2020{\natexlab{d}})Abdul-Mageed, Zhang, Hashemi,
  and Nagoudi}]{mageedL2020:aranet}
Muhammad Abdul-Mageed, Chiyu Zhang, Azadeh Hashemi, and El~Moatez~Billah
  Nagoudi. 2020{\natexlab{d}}.
\newblock {AraNet: A Deep Learning Toolkit for Arabic Social Media}.
\newblock In \emph{Proceedings of the 4th Workshop on Open-Source Arabic
  Corpora and Processing Tools, with a Shared Task on Offensive Language
  Detection}, pages 16--23.

\bibitem[{Al-Ayyoub et~al.(2018)Al-Ayyoub, Nuseir, Alsmearat, Jararweh, and
  Gupta}]{al2018deep}
Mahmoud Al-Ayyoub, Aya Nuseir, Kholoud Alsmearat, Yaser Jararweh, and Brij
  Gupta. 2018.
\newblock Deep learning for arabic nlp: A survey.
\newblock \emph{Journal of computational science}, 26:522--531.

\bibitem[{Al-Smadi et~al.(2020)Al-Smadi, Al-Zboon, Jararweh, and
  Juola}]{al2020transfer}
Mohammad Al-Smadi, Saad Al-Zboon, Yaser Jararweh, and Patrick Juola. 2020.
\newblock Transfer learning for {Arabic} named entity recognition with deep
  neural networks.
\newblock \emph{IEEE Access}, 8:37736--37745.

\bibitem[{Alharbi et~al.(2018)Alharbi, Magdy, Darwish, AbdelAli, and
  Mubarak}]{alharbi2018part}
Randah Alharbi, Walid Magdy, Kareem Darwish, Ahmed AbdelAli, and Hamdy Mubarak.
  2018.
\newblock Part-of-speech tagging for {Arabic} gulf dialect using bi-lstm.
\newblock In \emph{Proceedings of the Eleventh International Conference on
  Language Resources and Evaluation (LREC 2018)}.

\bibitem[{Alshehri et~al.(2020)Alshehri, Nagoudi, and
  Abdul-Mageed}]{alshehri2020osact4}
Ali Alshehri, El~Moatez~Billah Nagoudi, and Muhammad Abdul-Mageed. 2020.
\newblock \href {https://www.aclweb.org/anthology/2020.osact-1.6}
  {Understanding and detecting dangerous speech in social media}.
\newblock In \emph{Proceedings of the 4th Workshop on Open-Source Arabic
  Corpora and Processing Tools, with a Shared Task on Offensive Language
  Detection}, pages 40--47, Marseille, France. European Language Resource
  Association.

\bibitem[{Antoun et~al.(2020)Antoun, Baly, and Hajj}]{antoun2020arabert}
Wissam Antoun, Fady Baly, and Hazem Hajj. 2020.
\newblock Arabert: Transformer-based model for {Arabic} language understanding.
\newblock \emph{arXiv preprint arXiv:2003.00104}.

\bibitem[{Benajiba et~al.(2007)Benajiba, Rosso, and
  Bened$\backslash$'$\backslash$iruiz}]{Benajiba2007}
Yassine Benajiba, Paolo Rosso, and Jos{\'{e}}~Miguel
  Bened$\backslash$'$\backslash$iruiz. 2007.
\newblock {Anersys: An Arabic named entity recognition system based on maximum
  entropy}.
\newblock In \emph{International Conference on Intelligent Text Processing and
  Computational Linguistics}, pages 143--153.

\bibitem[{Bengio et~al.(2003)Bengio, Ducharme, Vincent, and
  Jauvin}]{bengio2003neural}
Yoshua Bengio, R{\'e}jean Ducharme, Pascal Vincent, and Christian Jauvin. 2003.
\newblock A neural probabilistic language model.
\newblock \emph{Journal of machine learning research}, 3(Feb):1137--1155.

\bibitem[{Bouamor et~al.(2019)Bouamor, Hassan, and Habash}]{bouamor2019madar}
Houda Bouamor, Sabit Hassan, and Nizar Habash. 2019.
\newblock {The MADAR shared task on Arabic fine{-}grained dialect
  identification}.
\newblock In \emph{{Proceedings of the Fourth Arabic Natural Language
  Processing Workshop}}, pages 199--207.

\bibitem[{Clark et~al.(2020)Clark, Luong, Le, and Manning}]{clark2020electra}
Kevin Clark, Minh-Thang Luong, Quoc~V Le, and Christopher~D Manning. 2020.
\newblock Electra: Pre-training text encoders as discriminators rather than
  generators.
\newblock \emph{arXiv preprint arXiv:2003.10555}.

\bibitem[{Conneau et~al.(2019)Conneau, Khandelwal, Goyal, Chaudhary, Wenzek,
  Guzm{\'a}n, Grave, Ott, Zettlemoyer, and Stoyanov}]{conneau2019unsupervised}
Alexis Conneau, Kartikay Khandelwal, Naman Goyal, Vishrav Chaudhary, Guillaume
  Wenzek, Francisco Guzm{\'a}n, Edouard Grave, Myle Ott, Luke Zettlemoyer, and
  Veselin Stoyanov. 2019.
\newblock Unsupervised cross-lingual representation learning at scale.
\newblock \emph{arXiv preprint arXiv:1911.02116}.

\bibitem[{Danet and Herring(2007)}]{danet2007multilingual}
Brenda Danet and Susan~C Herring. 2007.
\newblock \emph{The multilingual Internet: Language, culture, and communication
  online}.
\newblock Oxford University Press.

\bibitem[{Darwish(2013)}]{darwish2013named}
Kareem Darwish. 2013.
\newblock {Named entity recognition using cross-lingual resources: Arabic as an
  example}.
\newblock In \emph{Proceedings of the 51st Annual Meeting of the Association
  for Computational Linguistics (Volume 1: Long Papers)}, volume~1, pages
  1558--1567.

\bibitem[{Darwish et~al.(2018)Darwish, Mubarak, Abdelali, Eldesouki, Samih,
  Alharbi, Attia, Magdy, and Kallmeyer}]{darwish2018multi}
Kareem Darwish, Hamdy Mubarak, Ahmed Abdelali, Mohamed Eldesouki, Younes Samih,
  Randah Alharbi, Mohammed Attia, Walid Magdy, and Laura Kallmeyer. 2018.
\newblock Multi-dialect arabic pos tagging: a crf approach.
\newblock In \emph{Proceedings of the Eleventh International Conference on
  Language Resources and Evaluation (LREC 2018)}.

\bibitem[{Devlin et~al.(2018)Devlin, Chang, Lee, and
  Toutanova}]{devlin2018bert}
Jacob Devlin, Ming-Wei Chang, Kenton Lee, and Kristina Toutanova. 2018.
\newblock Bert: Pre-training of deep bidirectional transformers for language
  understanding.
\newblock \emph{arXiv preprint arXiv:1810.04805}.

\bibitem[{Dong and de~Melo(2019)}]{dong2019robust}
Xin~Luna Dong and Gerard de~Melo. 2019.
\newblock A robust self-learning framework for cross-lingual text
  classification.
\newblock In \emph{Proceedings of the 2019 Conference on Empirical Methods in
  Natural Language Processing and the 9th International Joint Conference on
  Natural Language Processing (EMNLP-IJCNLP)}, pages 6307--6311.

\bibitem[{El~Bazi and Laachfoubi(2019)}]{el2019arabic}
Ismail El~Bazi and Nabil Laachfoubi. 2019.
\newblock {Arabic} named entity recognition using deep learning approach.
\newblock \emph{International Journal of Electrical \& Computer Engineering
  (2088-8708)}, 9(3).

\bibitem[{Elaraby and Abdul-Mageed(2018)}]{elaraby2018deep}
Mohamed Elaraby and Muhammad Abdul-Mageed. 2018.
\newblock Deep models for {Arabic} dialect identification on benchmarked data.
\newblock In \emph{Proceedings of the Fifth Workshop on NLP for Similar
  Languages, Varieties and Dialects (VarDial 2018)}, pages 263--274.

\bibitem[{Gridach(2016)}]{gridach2016character}
Mourad Gridach. 2016.
\newblock Character-aware neural networks for {Arabic} named entity recognition
  for social media.
\newblock In \emph{Proceedings of the 6th workshop on South and Southeast Asian
  natural language processing (WSSANLP2016)}, pages 23--32.

\bibitem[{Hajmohammadi et~al.(2015)Hajmohammadi, Ibrahim, Selamat, and
  Fujita}]{hajmohammadi2015combination}
Mohammad~Sadegh Hajmohammadi, Roliana Ibrahim, Ali Selamat, and Hamido Fujita.
  2015.
\newblock Combination of active learning and self-training for cross-lingual
  sentiment classification with density analysis of unlabelled samples.
\newblock \emph{Information sciences}, 317:67--77.

\bibitem[{Herring et~al.(2015)Herring, Androutsopoulos
  et~al.}]{herring2015computer}
Susan~C Herring, Jannis Androutsopoulos, et~al. 2015.
\newblock Computer-mediated discourse 2.0.
\newblock \emph{The handbook of discourse analysis}, 2:127--151.

\bibitem[{Khalifa and Shaalan(2019)}]{khalifa2019character}
Muhammad Khalifa and Khaled Shaalan. 2019.
\newblock Character convolutions for arabic named entity recognition with long
  short-term memory networks.
\newblock \emph{Computer Speech \& Language}, 58:335--346.

\bibitem[{Kim et~al.(2017)Kim, Kim, Sarikaya, and
  Fosler-Lussier}]{kim2017cross}
Joo-Kyung Kim, Young-Bum Kim, Ruhi Sarikaya, and Eric Fosler-Lussier. 2017.
\newblock Cross-lingual transfer learning for pos tagging without cross-lingual
  resources.
\newblock In \emph{Proceedings of the 2017 conference on empirical methods in
  natural language processing}, pages 2832--2838.

\bibitem[{Kozareva et~al.(2005)Kozareva, Bonev, and Montoyo}]{kozareva2005self}
Zornitsa Kozareva, Boyan Bonev, and Andres Montoyo. 2005.
\newblock Self-training and co-training applied to spanish named entity
  recognition.
\newblock In \emph{Mexican International conference on Artificial
  Intelligence}, pages 770--779. Springer.

\bibitem[{Lample et~al.(2016)Lample, Ballesteros, Subramanian, Kawakami, and
  Dyer}]{lample2016neural}
Guillaume Lample, Miguel Ballesteros, Sandeep Subramanian, Kazuya Kawakami, and
  Chris Dyer. 2016.
\newblock {Neural architectures for named entity recognition}.
\newblock \emph{arXiv preprint arXiv:1603.01360}.

\bibitem[{Lample and Conneau(2019)}]{lample2019cross}
Guillaume Lample and Alexis Conneau. 2019.
\newblock Cross-lingual language model pretraining.
\newblock \emph{arXiv preprint arXiv:1901.07291}.

\bibitem[{Liu et~al.(2019)Liu, Ott, Goyal, Du, Joshi, Chen, Levy, Lewis,
  Zettlemoyer, and Stoyanov}]{liu2019roberta}
Yinhan Liu, Myle Ott, Naman Goyal, Jingfei Du, Mandar Joshi, Danqi Chen, Omer
  Levy, Mike Lewis, Luke Zettlemoyer, and Veselin Stoyanov. 2019.
\newblock Roberta: A robustly optimized bert pretraining approach.
\newblock \emph{arXiv preprint arXiv:1907.11692}.

\bibitem[{Ma and Hovy(2016)}]{ma2016end}
Xuezhe Ma and Eduard Hovy. 2016.
\newblock {End-to-end sequence labeling via bi-directional lstm-cnns-crf}.
\newblock \emph{arXiv preprint arXiv:1603.01354}.

\bibitem[{Maamouri et~al.(2004)Maamouri, Bies, Buckwalter, and
  Mekki}]{maamouri2004penn}
Mohamed Maamouri, Ann Bies, Tim Buckwalter, and Wigdan Mekki. 2004.
\newblock The penn {Arabic} treebank: Building a large-scale annotated {Arabic}
  corpus.
\newblock In \emph{NEMLAR conference on Arabic language resources and tools},
  volume~27, pages 466--467. Cairo.

\bibitem[{Maamouri et~al.(2014)Maamouri, Bies, Kulick, Ciul, Habash, and
  Eskander}]{maamouri2014developing}
Mohamed Maamouri, Ann Bies, Seth Kulick, Michael Ciul, Nizar Habash, and Ramy
  Eskander. 2014.
\newblock Developing an egyptian {Arabic} treebank: Impact of dialectal
  morphology on annotation and tool development.
\newblock In \emph{LREC}, pages 2348--2354.

\bibitem[{Mikolov et~al.(2013)Mikolov, Chen, Corrado, and
  Dean}]{mikolov2013efficient}
Tomas Mikolov, Kai Chen, Greg Corrado, and Jeffrey Dean. 2013.
\newblock Efficient estimation of word representations in vector space.
\newblock \emph{arXiv preprint arXiv:1301.3781}.

\bibitem[{Mitchell et~al.(2003)Mitchell, Strassel, Przybocki, Davis,
  Doddington, Grishman, Meyers, Brunstain, Ferro, and
  Sundheim}]{mitchell2003tides}
Alexis Mitchell, Stephanie Strassel, Mark Przybocki, J~Davis, George
  Doddington, Ralph Grishman, Adam Meyers, Ada Brunstain, Lisa Ferro, and Beth
  Sundheim. 2003.
\newblock Tides extraction (ace) 2003 multilingual training data.
\newblock \emph{LDC2004T09, Philadelphia, Penn.: Linguistic Data Consortium}.

\bibitem[{Obeid et~al.(2020)Obeid, Zalmout, Khalifa, Taji, Oudah, Alhafni,
  Inoue, Eryani, Erdmann, and Habash}]{obeid2020camel}
Ossama Obeid, Nasser Zalmout, Salam Khalifa, Dima Taji, Mai Oudah, Bashar
  Alhafni, Go~Inoue, Fadhl Eryani, Alexander Erdmann, and Nizar Habash. 2020.
\newblock {CAMeL Tools: An Open Source Python Toolkit for Arabic Natural
  Language Processing}.
\newblock In \emph{Proceedings of The 12th Language Resources and Evaluation
  Conference}, pages 7022--7032.

\bibitem[{Pan et~al.(2017)Pan, Zhang, May, Nothman, Knight, and
  Ji}]{pan2017cross}
Xiaoman Pan, Boliang Zhang, Jonathan May, Joel Nothman, Kevin Knight, and Heng
  Ji. 2017.
\newblock Cross-lingual name tagging and linking for 282 languages.
\newblock In \emph{Proceedings of the 55th Annual Meeting of the Association
  for Computational Linguistics (Volume 1: Long Papers)}, pages 1946--1958.

\bibitem[{Sagae(2010)}]{sagae2010self}
Kenji Sagae. 2010.
\newblock Self-training without reranking for parser domain adaptation and its
  impact on semantic role labeling.
\newblock In \emph{Proceedings of the 2010 Workshop on Domain Adaptation for
  Natural Language Processing}, pages 37--44.

\bibitem[{Shaalan(2014)}]{shaalan2014survey}
Khaled Shaalan. 2014.
\newblock A survey of {Arabic} named entity recognition and classification.
\newblock \emph{Computational Linguistics}, 40(2):469--510.

\bibitem[{Shaalan and Oudah(2014)}]{shaalan2014hybrid}
Khaled Shaalan and Mai Oudah. 2014.
\newblock A hybrid approach to {Arabic} named entity recognition.
\newblock \emph{Journal of Information Science}, 40(1):67--87.

\bibitem[{Shoufan and Alameri(2015)}]{shoufan2015natural}
Abdulhadi Shoufan and Sumaya Alameri. 2015.
\newblock Natural language processing for dialectical arabic: A survey.
\newblock In \emph{Proceedings of the second workshop on Arabic natural
  language processing}, pages 36--48.

\bibitem[{Song et~al.(2019)Song, Tan, Qin, Lu, and Liu}]{song2019mass}
Kaitao Song, Xu~Tan, Tao Qin, Jianfeng Lu, and Tie-Yan Liu. 2019.
\newblock Mass: Masked sequence to sequence pre-training for language
  generation.
\newblock \emph{arXiv preprint arXiv:1905.02450}.

\bibitem[{Tsai et~al.(2019)Tsai, Riesa, Johnson, Arivazhagan, Li, and
  Archer}]{tsai2019small}
Henry Tsai, Jason Riesa, Melvin Johnson, Naveen Arivazhagan, Xin Li, and Amelia
  Archer. 2019.
\newblock Small and practical bert models for sequence labeling.
\newblock \emph{arXiv preprint arXiv:1909.00100}.

\bibitem[{Van~Asch and Daelemans(2016)}]{van2016predicting}
Vincent Van~Asch and Walter Daelemans. 2016.
\newblock Predicting the effectiveness of self-training: Application to
  sentiment classification.
\newblock \emph{arXiv preprint arXiv:1601.03288}.

\bibitem[{Vaswani et~al.(2017)Vaswani, Shazeer, Parmar, Uszkoreit, Jones,
  Gomez, Kaiser, and Polosukhin}]{vaswani2017attention}
Ashish Vaswani, Noam Shazeer, Niki Parmar, Jakob Uszkoreit, Llion Jones,
  Aidan~N Gomez, {\L}ukasz Kaiser, and Illia Polosukhin. 2017.
\newblock Attention is all you need.
\newblock In \emph{Advances in neural information processing systems}, pages
  5998--6008.

\bibitem[{Wang et~al.(2007)Wang, Huang, and Harper}]{wang2007semi}
Wen Wang, Zhongqiang Huang, and Mary Harper. 2007.
\newblock Semi-supervised learning for part-of-speech tagging of mandarin
  transcribed speech.
\newblock In \emph{2007 IEEE International Conference on Acoustics, Speech and
  Signal Processing-ICASSP'07}, volume~4, pages IV--137. IEEE.

\bibitem[{Xie et~al.(2018)Xie, Yang, Neubig, Smith, and
  Carbonell}]{xie2018neural}
Jiateng Xie, Zhilin Yang, Graham Neubig, Noah~A Smith, and Jaime Carbonell.
  2018.
\newblock Neural cross-lingual named entity recognition with minimal resources.
\newblock \emph{arXiv preprint arXiv:1808.09861}.

\bibitem[{Xu and Rudnicky(2000)}]{xu2000can}
Wei Xu and Alex Rudnicky. 2000.
\newblock Can artificial neural networks learn language models?
\newblock In \emph{Sixth international conference on spoken language
  processing}.

\bibitem[{Yang et~al.(2019)Yang, Wang, Liu, Liu, Lyu, Wu, She, and
  Li}]{yang2019enhancing}
An~Yang, Quan Wang, Jing Liu, Kai Liu, Yajuan Lyu, Hua Wu, Qiaoqiao She, and
  Sujian Li. 2019.
\newblock Enhancing pre-trained language representations with rich knowledge
  for machine reading comprehension.
\newblock In \emph{Proceedings of the 57th Annual Meeting of the Association
  for Computational Linguistics}, pages 2346--2357.

\bibitem[{Zaidan and Callison-Burch(2011)}]{zaidan2011arabic}
Omar Zaidan and Chris Callison-Burch. 2011.
\newblock The arabic online commentary dataset: an annotated dataset of
  informal arabic with high dialectal content.
\newblock In \emph{Proceedings of the 49th Annual Meeting of the Association
  for Computational Linguistics: Human Language Technologies}, pages 37--41.

\bibitem[{Zaidan and Callison-Burch(2014)}]{zaidan2014arabic}
Omar~F Zaidan and Chris Callison-Burch. 2014.
\newblock Arabic dialect identification.
\newblock \emph{Computational Linguistics}, 40(1):171--202.

\end{thebibliography}
\bibliographystyle{acl_natbib}

\newpage
\clearpage
\appendix
\section*{Appendices} 




  

\section{POS Tag Set}
\label{sec:pos-tagset}
Table ~\ref{Tab:pos-tags} lists all the part-of-speech (POS) tags used in our experiments.
\begin{table}[h!]
\begin{center}
\begin{tabular}{llll}
\toprule
\bf Tag &  \bf Description &  \bf Tag & \bf Description  \\ 
\hline
ADV & adverb & ADJ & adjective    \\
CONJ & conjunction & DET & determiner  \\
NOUN  & noun & NSUFF & noun suffix    \\
NUM   & number  & PART & particle  \\
PUNC & punctuation  & PRON & pronoun \\
PREP & preposition & V & verb   \\
ABBREV & abbreviation &  VSUFF  & verb suffix \\
FOREIGN & non-Arabic & FUT\_PART & future particle   \\
PROG\_PART & progressive particle & EMOT & Emoticon/Emoji\\
MENTION & twitter mention & HASH & Hashtag \\
URL & URL & ~~~ -- & ~~~ --\\
\bottomrule
\end{tabular}
\end{center}
\caption{\label{Tab:pos-tags} The POS tag set in~\cite{darwish2018multi}.}
\end{table}

\section{Error Analysis}\label{append:error}
\label{app-err}

\begin{table*}[h!]
 \footnotesize
\setlength{\extrarowheight}{2pt}
\begin{tabular}{cc|c|c|c|c|}
  & \multicolumn{1}{c}{} & \multicolumn{4}{c}{Predicted} \\
  & \multicolumn{1}{c}{} & \multicolumn{1}{c}{PER}  & \multicolumn{1}{c}{LOC}  & \multicolumn{1}{c}{ORG} & \multicolumn{1}{c}{O} \\\cline{3-6}
            & PER & 117 & 2 & 2 & 66 \\ \cline{3-6}
Gold  & LOC & 11 & 33 & 1 & 39\\\cline{3-6}
            & ORG & 5 & 5 & 5 & 57\\\cline{3-6}
            & O & 130 & 14 & 15 & 5,940 \\\cline{3-6}

\end{tabular}
\begin{tabular}{cc|c|c|c|c|}
  & \multicolumn{1}{c}{} & \multicolumn{4}{c}{Predicted} \\
  & \multicolumn{1}{c}{} & \multicolumn{1}{c}{PER}  & \multicolumn{1}{c}{LOC}  & \multicolumn{1}{c}{ORG} & \multicolumn{1}{c}{O} \\\cline{3-6}
            & PER & 120 & 3 & 2 & 62 \\ \cline{3-6}
Gold  & LOC & 10 & 34 & 0 & 40\\\cline{3-6}
            & ORG & 5 & 6 & 11 & 66\\\cline{3-6}
            & O & 54 & 8 & 2 & 6,035 \\\cline{3-6}

\end{tabular}
\caption{NER confusion matrices for fine-tuning (left) and self-training (right) on the development set of the DA NER data.}
\label{tab:conf-mat}
\end{table*}

\begin{table*}[h!]
    \centering
    \footnotesize 

    \begin{tabular}{l|c|l|c|l|c}
        \toprule
        \textbf{no.} & \textbf{Token} &\textbf{Eng.} &\textbf{MSA} & \textbf{Context/Explanation} & \textbf{FT Pred.} \\
        \hline
        (1) & \<نبي > & we want  & \<نريد>  &  \< نبي نعرف من...>    (\textit{we want to know who}) & PER  \\

       (2) &   \<ماكانوا> & wasn't &  \<لم يكونوا>
          & \<أغلب الي ماكانوا مصدقين>
          (\textit{most of those who wasn't believing}) & LOC  \\

         (3) & \<لوووول > & LOL &  \<ضحك>  & \< لوووول...> (interjection) & PER  \\
        (4) & \< عشان > & for & \<لكي > &\<تبي بطاريات عشان تلعب > (\textit{she wants batteries to play})  & LOC  \\

       (5) &  \<دلوقتي > & now &  \<الآن >  & \<...اقنعوه ينزل دلوقتي>  \textit{(convince him to move now})  &         PER  \\

       (6) &   \<ايش> & what &  \<ماذا>
          & \<ايش رأيك>
          (\textit{what do you think?}) & PER \\

        (7) & \<قادر > & capable &  \<قادر> &\<وبقدرة قادر...>  (\textit{magically}; idiomatic expression)& PER \\

       (8) &  \<المشين > & shameful &  \<المشين >  & \<المشين طنطاوي >  (\textit{shameful Tantawy}; Playful for \textit{General Tant.})& PER \\

       (9) &  \<ايديكوا > & your hands  &  \<أيديكم >   &   \<ابوس ايديكوا اقنعوه... >   \textit{(I entreat you to convince hi}m) & PER \\

      (10) &  \<اسالك > & I ask you &  \<أسألك> & \< ودي اسالك شنهي > (\textit{I ask you what}) & ORG  \\
      (11) &   \<مين > & who &  \<مين>
          & \<صوتك مع مين البدوي >
          (\textit{who do you vote for, Badawi}) & PER  \\
        
       (12) &   \<فلوبي ديسك> & floppy disk &  \<قرص مرن>  & \<ماسك عليه فلوبي ديسك >  (\textit{holds a floppy disk against him}) & PER  \\

       (13) &   \<لحبايب> & loved ones &  \<الأحباء> & \<تعال علم يف لحبايب > (\textit{come teach your loved ones}) & LOC \\
          
        (14) & \<ماي> & water & \<ماء> & \<جبت لهم ماي> \textit{(brought them water)} & PER  \\

    (15) & \<ريتويت> & retweet & \<إعادة تغريد> & \<لو قرفان دوس  ريتويت> \textit{(if depressed click retweet)} & PER \\
    
   \bottomrule
    \end{tabular}
    \caption{\textbf{NER task.} Bigger sample false positives mitigated by self-training. These were correctly predicted as the unnamed entity ``O" by the self-trained model.}
    \label{tab:fp_analysis_app}
\end{table*}


\begin{table*}[h!]
    \footnotesize 
    \centering

\end{table*}

\begin{table*}[h!]
    \centering
    \footnotesize
     \begin{tabular}{l|c|l|c|c|c}
        \toprule
        \textbf{no.} & \textbf{Token(s)} & \textbf{Context/Explanation} & \textbf{Gold} & \textbf{FT} & \textbf{ST}  \\
        \hline
        (1) & \<بالمبارك> & \<بالمبارك عاد احنا> \textit{(We are still in Mubarak}) & LOC & PER & O \\
        (2) & \<محشش> &   \<محشش دخل المحاضرة>
        \textit{(a drunk entered the lecture)} &  O & PER & PER \\
        (3) & \<بأمارة> & \<بأمارة ايه وفين>
        \textit{(what is the evidence/sign and where?)} &  O & LOC & LOC \\
        (4) & \<لمستفشي> & \<لمستفشي قصر الدوباره>
        \textit{(to Qasr AlDobara Hospital)} & LOC & O & O \\
        (5) & \<كنتاكي> & \<عند كنتاكي>
        \textit{(by Kentucky [resturant])} & LOC & O & O \\
        (6) & \<داون تاون> & \<مشروع داون تاون بطنطا>
        \textit{(a down town Tanta project)}&
        LOC & O & O \\
        (7) & \<يابطل> & \<مبروك يابطل> 
        \textit{(Congratulations, hero!)} &  O & PER & PER \\
        (8) & \<الاخوان> & \<نختلف مع الاخوان>  \textit{(we disagree with the Muslim brotherhood)} & ORG & O & O \\
        (9) & \<قناة العربية> & \<شفت قناة العربية>  \textit{(watched Al Arabya Channel)}&  ORG & O & O \\
        (10) & \<المجلس العسكري> &  \<اللي عمله المجلس العسكري> \textit{(what the military council did)} & ORG & O & O \\
         \bottomrule
    \end{tabular}
    
    \caption{\textbf{NER task.} Sample errors that are not fixed by self-training (shared with the mere fine-tuned model)}
    \label{tab:shared-errors}
\end{table*}

\end{document}